%% file: main.tex
\documentclass[10pt,twocolumn,letterpaper]{article}

\usepackage{wacv}
\usepackage{times}
\usepackage{epsfig}
\usepackage{graphicx}
\usepackage{amsmath}
\usepackage{amssymb}
\usepackage{booktabs}
\usepackage[accsupp]{axessibility}  

\def\wacvPaperID{32} 

\wacvalgorithmstrack   

\wacvfinalcopy 

\ifwacvfinal
\usepackage[breaklinks=true,bookmarks=false]{hyperref}
\else
\usepackage[pagebackref=true,breaklinks=true,colorlinks,bookmarks=false]{hyperref}
\fi

\input{preamble}

\begin{document}

\title{
Toward
Edge-Efficient Dense Predictions
with
Synergistic Multi-Task 
Neural Architecture Search
}

\author{
\vspace{-1.5em}
\\ 
{
    Thanh Vu$^1$\thanks{Work done during an internship at X. \qquad $^\dag$Work done while at X.}
    \qquad
    Yanqi Zhou$^2$
    \qquad
    Chunfeng Wen$^{3\dag}$
    \qquad
    Yueqi Li$^3$
    \qquad
    Jan-Michael Frahm$^1$
}
\\ 
\\
{
    ~
    $^1$UNC at Chapel Hill
    ~~~\qquad
    $^2$Google Research
    ~~\qquad\quad
    $^3$X, The Moonshot Factory
    ~~\quad
}
\\ 
{\tt\small
    \{tvu,jmf\}@cs.unc.edu
    \qquad
    \{yanqiz\}@google.com
    \qquad
    \{fannywen,yueqili\}@google.com
}
}

\maketitle
\thispagestyle{empty}

\input{fig/teaser}
\input{sec/0_abstract}

\input{sec/1_introduction}
\input{sec/2_related}
\input{sec/3_method}
\input{sec/4_results}
\input{sec/5_conclusions}
\clearpage
{\small
\bibliographystyle{ieee_fullname}
\bibliography{main}
}
    
\input{sec/X_supplementary}


\end{document}

%% file: preamble.tex
\usepackage{color}
\usepackage{comment}
\usepackage{enumitem} 
\usepackage{overpic} 
\usepackage{booktabs}
\usepackage{textcomp,gensymb}
\usepackage{subcaption}
\usepackage{capt-of}
\usepackage{cuted}

\usepackage[capitalize]{cleveref}
\crefname{section}{Sec.}{Secs.}
\Crefname{section}{Section}{Sections}
\Crefname{table}{Table}{Tables}
\crefname{table}{Tab.}{Tabs.}

\definecolor{turquoise}{cmyk}{0.65,0,0.1,0.3}
\definecolor{purple}{rgb}{0.65,0,0.65}
\definecolor{dark_green}{rgb}{0, 0.5, 0}
\definecolor{orange}{rgb}{0.8, 0.6, 0.2}
\definecolor{red}{rgb}{0.8, 0.2, 0.2}
\definecolor{darkred}{rgb}{0.6, 0.1, 0.05}
\definecolor{blueish}{rgb}{0.0, 0.3, .6}
\definecolor{light_gray}{rgb}{0.7, 0.7, .7}
\definecolor{pink}{rgb}{1, 0, 1}
\definecolor{greyblue}{rgb}{0.25, 0.25, 1}

\newcommand{\Fig}[1]{Fig.~\ref{fig:#1}}

\newcommand{\Tab}[1]{Tab.~\ref{tab:#1}}
\newcommand{\Table}[1]{Table~\ref{tab:#1}}

\newcommand{\Eq}[1]{Eq.~\ref{eq:#1}}

\usepackage{blindtext}

\renewcommand{\paragraph}[1]{\vspace{1em}\noindent\textbf{#1}.}

\usepackage{float}
\usepackage{placeins}
\usepackage{wrapfig}

%% file: fig/teaser.tex
\begin{strip} 
\vspace{-5em}
\begin{center}
\captionsetup{type=figure}
\includegraphics[width=0.37\linewidth]{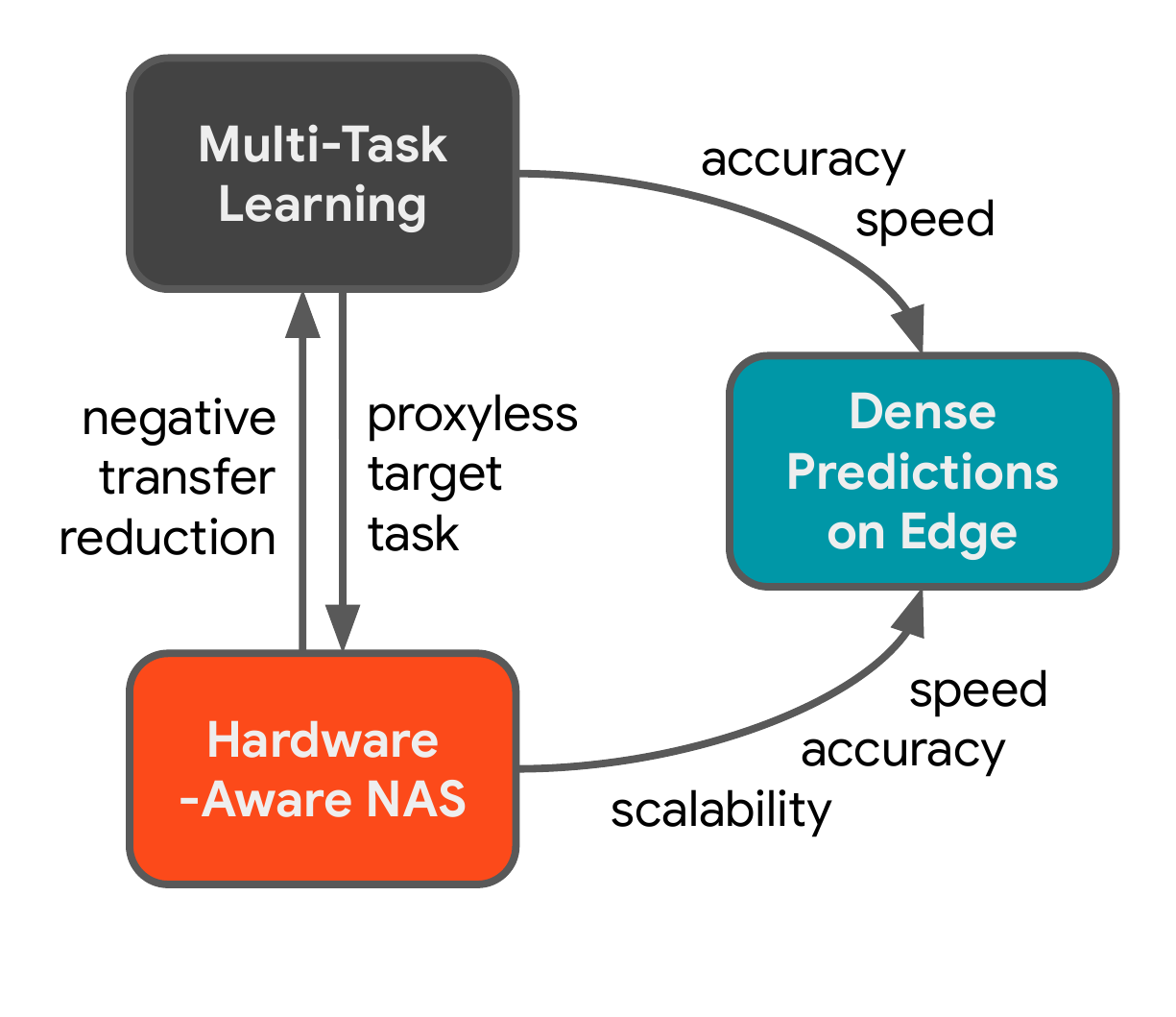}
\hspace{3em}
\includegraphics[width=0.47\linewidth]{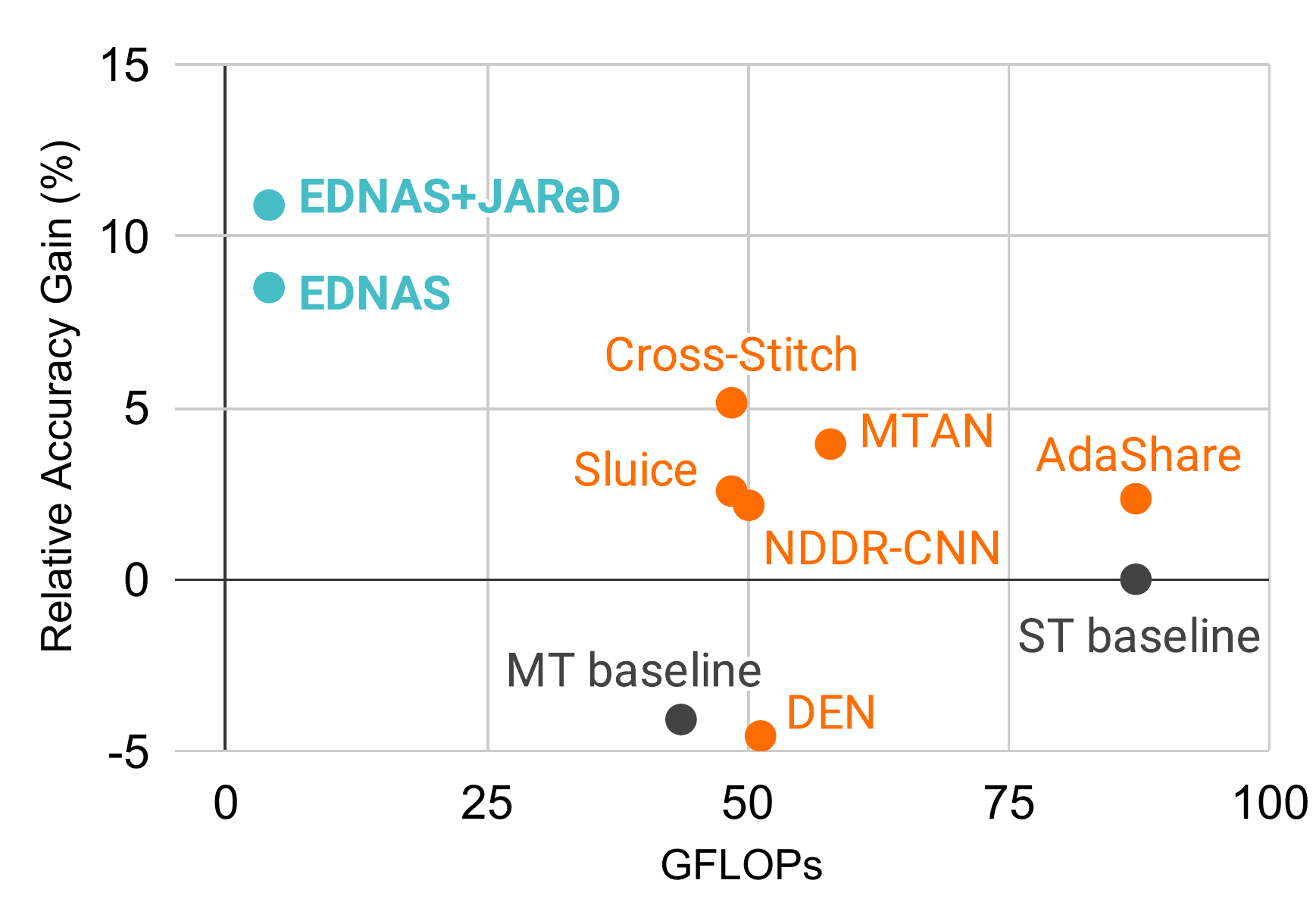}
\captionof{figure}{
    \textbf{An overview of our proposed methods}. 
    First, EDNAS framework leverages the synergy and joint learning of multi-task dense prediction (MT-DP) and hardware-aware NAS to both complement each component and boost on-device performance.
    On the left is an illustration of the synergistic relationship of these components.
    Second, JAReD loss reduces depth estimation noise and further improves accuracy.
    On the right is the performance of our proposed techniques on CityScapes compared to state-of-the-art MT-DP approaches. 
}
\label{fig:teaser}
\end{center}
\end{strip}


%% file: sec/0_abstract.tex
\begin{abstract}
\vspace{-1em}
In this work, we propose a novel and scalable solution to address the challenges of developing efficient dense predictions on edge platforms. 
Our first key insight is that Multi-Task Learning (MTL) and hardware-aware Neural Architecture Search (NAS) can work in synergy to greatly benefit on-device Dense Predictions (DP). 
Empirical results reveal that the joint learning of the two paradigms is surprisingly effective at improving DP accuracy, 
achieving superior performance over both the transfer learning of single-task NAS and prior state-of-the-art approaches in MTL, all with just 1/10th of the computation. 
To the best of our knowledge, our framework, named EDNAS, is the first to successfully leverage the synergistic relationship of NAS and MTL for DP. 
Our second key insight is that the standard depth training for multi-task DP can cause significant instability and noise to MTL evaluation. 
Instead, we propose JAReD, an improved, easy-to-adopt Joint Absolute-Relative Depth loss, that reduces up to 88\% of the undesired noise while simultaneously boosting accuracy. 
We conduct extensive evaluations on standard datasets, benchmark against strong baselines and state-of-the-art approaches, as well as provide an analysis of the discovered optimal architectures.
\end{abstract}

%% file: sec/1_introduction.tex
\vspace{-1em}
\section{Introduction}
\label{sec:intro}
Recent years have witnessed a strong integration of computer vision in many downstream edge applications such as autonomous driving \cite{price_isorc20,simple_det,self_supervised_pillar_motion,fusion_transformer,waymo_open,rethink_ming19,cross_modality_3d}, mobile vision~\cite{chamnet,mobilenetv3,mobilenets,awn,building_rec_ncur16,fbnet}, robotics \cite{rt_seg_fast_attention,yolactedge,robotics}, and even computational agriculture \cite{Chiu_2020_CVPR,Kamilaris2018DeepLI,robot_cabbage}, fueled by rapid innovations of deep neural networks.
In many of these applications, pixel-level dense prediction tasks such as semantic segmentation or depth estimation can play a critical role.
For example, self-driving agents are using semantic and depth information to detect lanes, avoid obstacles, and locate their own positions.
In precision agriculture, the output of these tasks can be used for crop analysis, yield prediction, in-field robot navigation, etc. As more and more neural models are being deployed into the real world, there has been a continuously growing interest in developing edge-efficient architectures for dense predictions over the years.

However, designing fast and efficient dense prediction models for edge devices is challenging.
First of all, pixel-level predictions such as semantic segmentation and depth estimation are fundamentally slower than some other popular vision tasks, including image classification or object detection.
This is because after encoding the input images into low-spatial resolution features, these networks need to upsample them back to produce high-resolution output masks.
In fact, dense estimation can be several times or even an order of magnitude slower than their counterparts, depending on the specific model, hardware, and target resolution.
Thus, real-time dense prediction models are not only non-trivial to design, they can easily become a latency bottleneck in systems that utilize their outputs.
Such problems are intensified for edge applications on  platforms like the Coral TPU \cite{coral_edgetpu} due to the limited computational resources, despite the need for low latency, e.g., to inform the users or process subsequent tasks in real time.

Second, developing models for these edge environments is costly and hard to scale in practice. 
On one hand, the architectural design process requires a significant amount of time, human labor, and expertise, with the development process ranging from a few months to a couple of years.
On the other hand, edge applications may require deployment on various platforms, including cell phones, robots, drones, and more. Unfortunately, optimal designs discovered for one hardware may not generalize to another.
All of these together pose challenges to the development of fast and efficient models for on-edge dense predictions.

To tackle these problems,
our first key insight is that Multi-Task Learning of Dense Predictions (MTL-DP or MT-DP) and hardware-aware Neural Architecture Search (h-NAS) can work in synergy to not only mutually benefit but also significantly improve accuracy and computation.
To the best of our knowledge, our framework, named
\hspace{1em}
\textit{EDNAS}\footnote{short for ``Edge-Efficient Dense Predictions via Multi-Task NAS"}, is the first to successfully exploit such a synergistic relationship of NAS and MTL for dense predictions.
%
Indeed, on one hand, state-of-the-art methods for multi-task dense predictions \cite{bmtas,learning2branch,mtan,single_tasking_attentive,adashare,branched_mt,pcgrad},
in which related tasks are learned jointly together, 
mostly focus on learning \textit{how to share a fixed set} of model components effectively among tasks but do not consider if such a set itself is \textit{optimal} for MTL to begin with.
Moreover, these works typically study large models targeting powerful graphic accelerators such as V100 GPU for inference and are not readily suitable for edge applications. 
On the other hand, NAS methods aim to automatically learn an optimal set of neural components and their connections.
However, the current literature often focuses on either simpler tasks such as classification \cite{nas_mt_cls_multi_obj,nas_mt_cls_evol,nas_mt_cls_multipath} or single-task training setup \cite{nas_seg_densely,autodeeplab}.
In contrast, we jointly learn MTL-DP and NAS and leverage their strengths to tackle the aforementioned issues simultaneously, resulting in a novel and improved approach to efficient dense predictions for edge.
%

Our second key insight is that the standard depth estimation training used in MTL-DP
can produce significant fluctuation in the evaluation accuracy.
Indeed, our analysis reveals a potential for undesirably large variance in both absolute and relative depth.
We hypothesize that this is caused by the standard depth training practice that relies solely on $\mathcal{L}_1$ loss function.
This can significantly and negatively affect the accuracy of MT-DP evaluation as arbitrary ``improvement'' (or ``degradation") can manifest purely because of random fluctuation in the relative error.
It is important that we raise awareness of and appropriately address this issue as segmentation and depth information are arguably two of the most commonly jointly learned and used tasks in edge applications.
To this end, we propose \textit{JAReD}, an easy-to-adopt augmented loss that jointly and directly optimizes for both relative and absolute depth errors.
The proposed loss is highly effective at simultaneously reducing noisy fluctuations and boosting overall prediction accuracy.

We conduct extensive evaluations on CityScapes \cite{cityscapes} and NYUv2 \cite{nyuv2} to demonstrate the effectiveness and robustness of EDNAS and JAReD loss.
Experimental results indicate that our methods can yield significant gains, up to +8.5\% and +10.9\% DP accuracy respectively, considerably higher than the previous state of the art, with only 1/10th of the parameter and FLOP counts (\Fig{teaser}).

%% file: sec/2_related.tex
\section{Background and Related Works}
\label{sec:related}
\input{fig/overview}
In general, dense prediction models are often designed manually, in isolation, or not necessarily constrained by limited edge computation \cite{deeplabv1,rt_seg_fast_attention,autodeeplab,yolactedge}.
Specifically, works on multi-task learning for dense predictions (MTL-DP) \cite{bmtas,rel_context,nas_mt_soft_mtlnas,learning2branch,adashare,branched_mt} often take a fixed base architecture such as DeepLab \cite{deeplab_v2_2017} and focus on learning to effectively shared components, e.g. by cross-task communication modules \cite{rel_context,nas_mt_soft_mtlnas}, adaptive tree-like branching~\cite{bmtas,learning2branch,branched_mt},  layer skipping \cite{adashare}, etc. (\Fig{overview}).
On the other hand, neural architecture search (NAS) studies up until recently have focused mostly on either image classification problems\cite{nas_mt_cls_elastic,nas_mt_cls_multi_obj,nas_mt_cls_virtual,nas_mt_cls_evol,nas_mt_cls_snr,nas_mt_cls_multipath}
or learning tasks in isolation \cite{nas_seg_densely,autodeeplab,mnasnet,nahas}. Few have explored architecture search for joint training of dense prediction tasks.
However, as mentioned earlier, edge efficiency can potentially benefit both MTL-DP and NAS.
To the best of our knowledge, our study is the first to report successful joint optimization of these two learning paradigms for dense predictions.
Next, we give an overview of the most relevant efforts in the two domains of MTL and NAS.
For more details, please refer to these comprehensive surveys: MTL \cite{mlt_1997,survey_mt}, MTL for dense predictions \cite{survey_mt_dense}, NAS \cite{survey_nas}, and hardware-aware NAS \cite{survey_nas_hw}, .

\paragraph{Neural Architecture Search (NAS)} In the past few years, neural architecture search (NAS) has emerged as a solution to automate parts of the network design process. NAS methods have shown remarkable progress and outperformed many handcrafted models \cite{autodeeplab,mnasnet,efficientnetv1,efficientnetv2}. 
In our case, we are interested in hardware-aware NAS \cite{proxylessnas,fbnet,nahas} which can discover efficient architectures suitable for one or multiple targeted edge platforms.
This is typically done by casting hardware-aware NAS as a multi-objective optimization problem \cite{proxylessnas,mnasnet,fbnet} and adding hardware cost, e.g. latency, memory, and energy, alongside prediction accuracy, to guide the search.
However, current studies often focus on image classification \cite{nas_mt_cls_elastic,nas_mt_cls_multi_obj,nas_mt_cls_virtual,nas_mt_cls_evol,nas_mt_cls_snr,nas_mt_cls_multipath} 
or learning tasks in isolation \cite{mnasnet,nahas}. 
However, performing multiple dense prediction tasks simultaneously can have significant benefits for both inference speed and accuracy since tasks can leverage each other's training signals as inductive biases to improve their own learning and the model's generalization~\cite{mlt_1997}.
Thus, we are interested in combining hardware-aware NAS with multi-task learning of dense prediction tasks to achieve both better accuracy and better inference speed on edge devices.
To this end, there have been only a limited number of studies \cite{bmtas,learning2branch,adashare,branched_mt} that started to explore similar problems, which we will discuss next.

\paragraph{MTL for Dense Predictions}
The goal of Multi-Task Learning (MTL) \cite{mlt_1997,survey_mt} is to jointly learn multiple tasks together to leverage cross-task information to improve per-task prediction quality. 
In the context of edge applications, we are also interested in the property of MTL that lets tasks share computation and output multiple task predictions in one pass, thereby improving the overall inference speed. 
%
This 
is particularly useful for dense predictions because they tend to be more computationally expensive than their counterparts such as classification \cite{mobilenetv3,mobilenetv1,mobilenetv2,efficientnetv1,efficientnetv2} or detection \cite{efficientdet,mobiledet}.
A popular formulation of MTL that accomplishes this goal is called hard parameter sharing (HPS)~\cite{mtan,pcgrad}. Compared to soft parameter sharing (SPS)~\cite{nas_mt_soft_mtlnas}, whose multi-task model size scales linearly with the number of tasks due to separate per-task sub-networks, HPS models are more edge-friendly due to their compact architectural structure. 
Specifically, HPS architectures are typically composed of a shared trunk that extracts joint features for all tasks and multiple per-task heads or branches that take the extracted features as input and produce specific task prediction. The most standard setup is to have all task heads branch off at the same point \cite{mtan}. This is also our setup of choice for the scope of this work. 
In addition, recent studies have begun to explore strategies to learn adaptive sharing architectures from data \cite{bmtas,learning2branch,single_tasking_attentive,adashare,branched_mt}. Attention~\cite{single_tasking_attentive} and Layer-skipping\cite{adashare} have been used to efficiently learn a single shared model while modifying their behaviors to output the desired task-specific prediction, given a task. 
Other studies \cite{bmtas,learning2branch,branched_mt} opt to augment the HPS architectures by learning the branching of tasks. In other words, the learned models may have multiple splitting points, where some tasks can branch off earlier while some others share more layers.
A common theme of these approaches is that given a fixed starting architecture,
the focus is on learning \textit{which components} of such network should be shared.
Our work shifts the focus to the base network and instead asks \textit{what components} should be included in such architecture to best benefit multi-task dense predictions. 



%% file: fig/overview.tex
\newcommand{\imgsizeA}{0.23}
\newcommand{\imgsizeAspace}{0.07in}
\begin{figure*}[t]
\begin{center}
    \begin{subfigure}[b]{\imgsizeA\linewidth}
        \centering
        \captionsetup{justification=centering}
        \includegraphics[width=\linewidth]{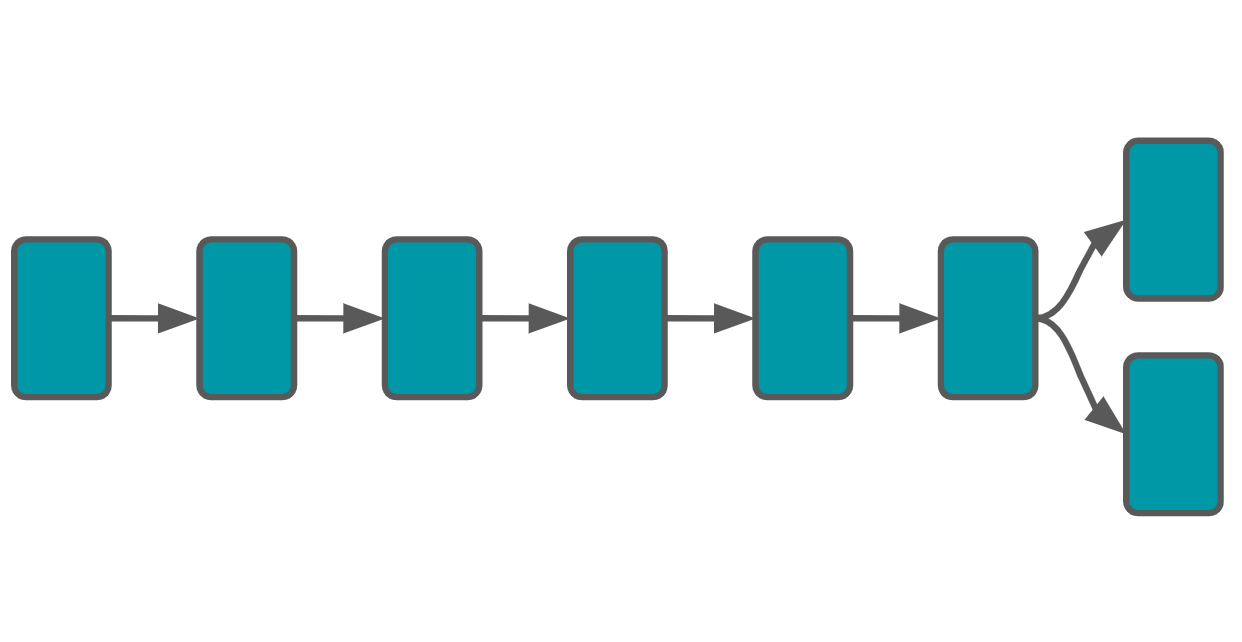}
        \caption*{{(a) Hard parameter sharing \cite{mtan,pcgrad}}}
    \end{subfigure}
    \hspace{\imgsizeAspace}
    \begin{subfigure}[b]{\imgsizeA\linewidth}
        \centering
        \captionsetup{justification=centering}
        \includegraphics[width=\linewidth]{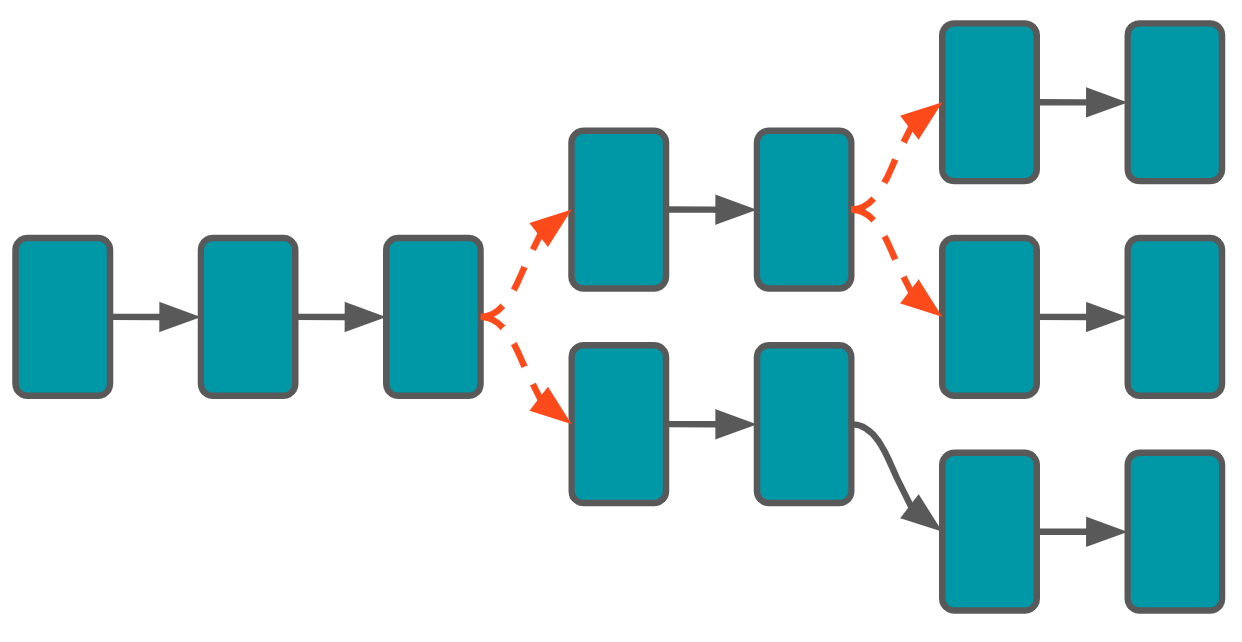}
        \caption*{{(b) Learning to branch \cite{learning2branch,bmtas,branched_mt}}}
    \end{subfigure}
    \hspace{\imgsizeAspace}
    \begin{subfigure}[b]{\imgsizeA\linewidth}
        \centering
        \captionsetup{justification=centering}
        \includegraphics[width=\linewidth]{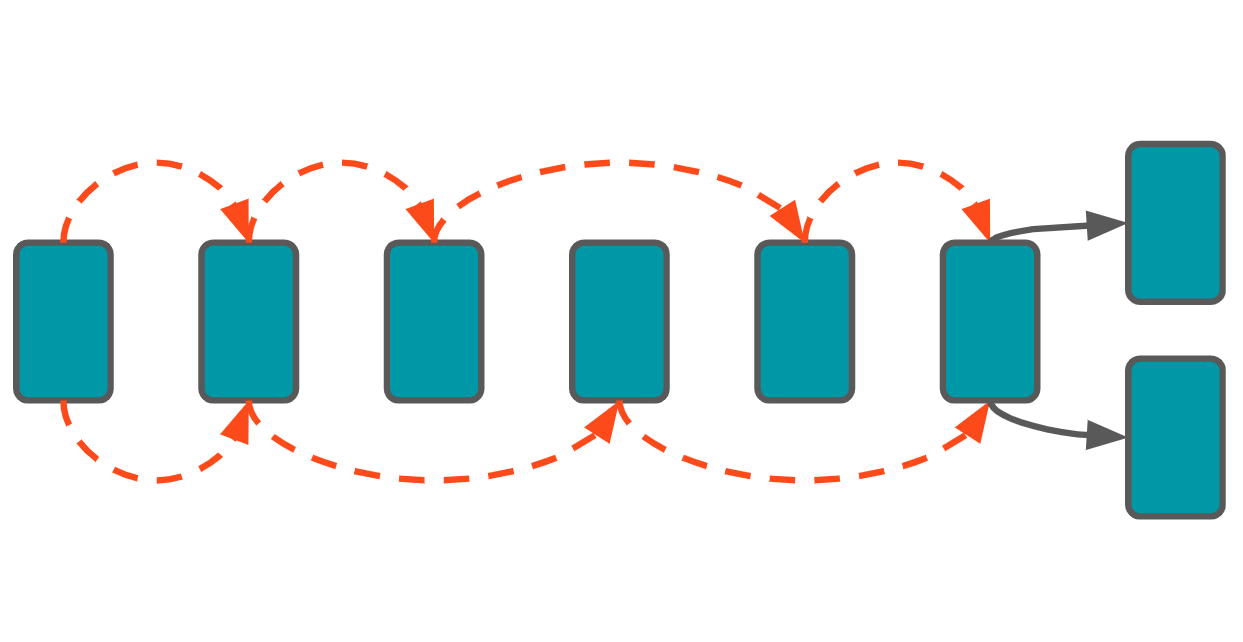} 
        \caption*{{(c) Learning to skip layers \cite{adashare}}}
    \end{subfigure}
    \hspace{\imgsizeAspace}
    \begin{subfigure}[b]{\imgsizeA\linewidth}
        \centering
        \captionsetup{justification=centering}
        \includegraphics[width=\linewidth]{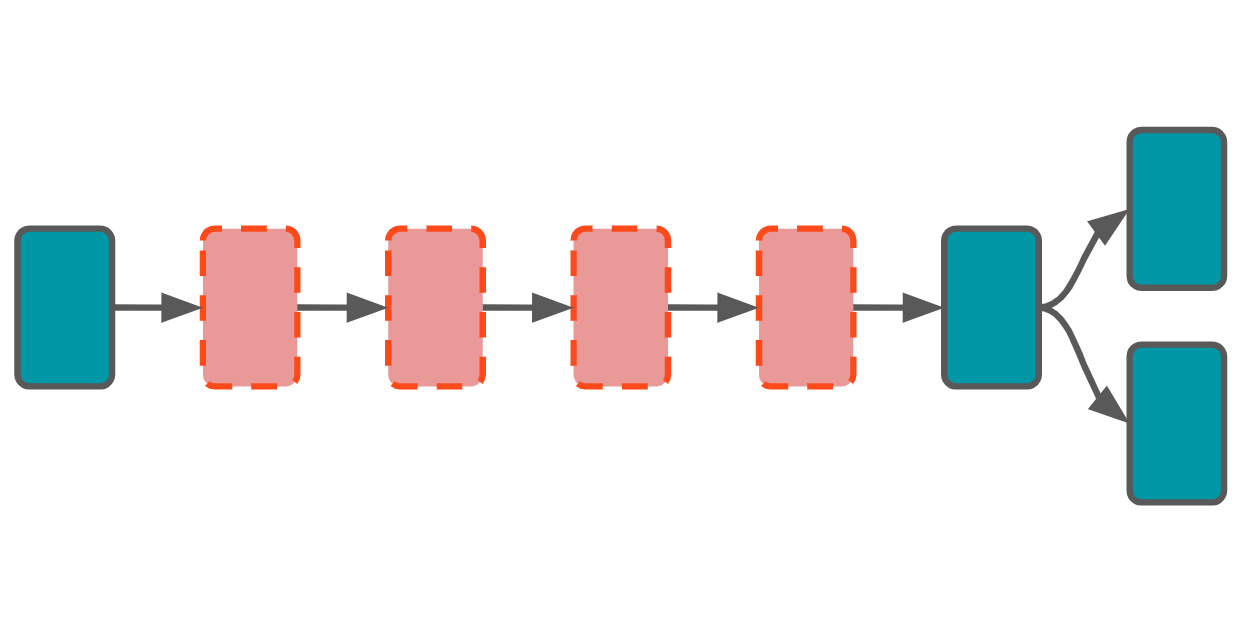} 
        \caption*{{(d) Searching for layers (ours)}}
    \end{subfigure}
\end{center}
\vspace{-0.1in}
\caption{
\textbf{Conceptual comparison with existing approaches.}
While current MT-DP methods focus on how to \textit{better share a fixed set of layers}, we instead learn \textit{better sets of layers to share}. Components in {\color{red}red} are learnable while others are fixed
}
\label{fig:overview}
\end{figure*}

%% file: sec/3_method.tex
\section{Methodology}
{
} 

\subsection{EDNAS: Joint MTL-DP and h-NAS}
\label{sec:nas}
\vspace{-0.5em}
\paragraph{Synergistic Joint Learning}
\label{sec:joint-learning}
Our key idea is that we can leverage multi-task inference to significantly reduce computation across several dense prediction tasks, while utilizing hardware-aware NAS to simultaneously improve edge latency, design scalability, and multi-task learning. Combining these two paradigms, MT-DP and NAS, is beneficial not only to edge inference but also to each other.
\Fig{teaser} illustrates these relationships.
First, regarding edge applications,
multi-task models \cite{survey_mt_dense} that output several predictions at once are attractive since they share computation across tasks to avoid multiple inference runs and improve the overall latency linearly by design.
However, this multi-task setup also leads to performance degradation, known as negative transfer.
While most current works attribute this problem to improper sharing of neural components, we hypothesize that
components of popular base networks such as DeepLab \cite{deeplab_v2_2017} -ResNet \cite{resnet} may be well-tuned for their original individual task, but not necessarily optimal for multi-task setting. 
It is possible that certain layers, for example, may need more channels to capture nuanced features required when the number of tasks increases.
Moreover, these models may need to be deployed on different edge platforms and thus, their components need to be optimized accordingly.
This motivates us to explore NAS as a systematic and scalable method to discover components that could be more suitable for multi-task learning and edge inference. 
%
Second, from the perspective of NAS, directly searching for multi-task architectures can potentially yield better results than transferring single-task searched architectures to multi-task settings post NAS.
In a way, we are removing a proxy target and its assumption that architectures, which are good for an individual task such as segmentation, are also optimal for multi-task learning. 


\paragraph{Hardware-Aware Multi-Task Objective}
Given a fixed set of $N$ tasks $T = \{T_1, T_2, ... T_N\}$, we formulate the problem of multi-task NAS as a multi-objective search. 
Our goal is to discover optimal models with both high accuracy for all tasks in $T$ and low inference latency on specific edge devices.
Let $a$ be an architecture with weights $w_a$ sampled from the search space $A$ and $h$ be a target edge hardware. Our optimization can then be expressed as follows:
\begin{equation}
  \max_{a \subset A} Rwd(a, T, h, w_a^*)
  \label{eq:optim_reward}
\end{equation}
\begin{equation}
  \text{s.t.} \quad 
  w_a^*=arg\min_{w_a} Loss(a, T, w_a) 
\end{equation}
\begin{equation}
  \text{and} \quad 
  Lat(a, h) \leq l_h
\end{equation}
with $Rwd()$ being the objective or reward function and $l_h$ being the target edge latency dependent on the hardware and application domain.
Inspired by \cite{mnasnet}, we use a weighted product for the reward function $Rwd()$ to jointly optimize for models' accuracy and latency constrained by hardware-dependent requirements such as inference latency, chip area, energy usage, etc. 
This allows for flexible customization and encourages Pareto optimal solutions of multi-objective learning \cite{pareto}.
In this work, we focus on inference latency $Lat(a,h)$ as the main hardware constraint.
\begin{equation}
    Rwd(a,T,h,w_a) = Acc(a,T,w_a) \left [\frac{Lat(a, h)}{l_h}\right]^{\beta}
    \label{eq:reward}
\end{equation}
\begin{equation}
    \text{s.t.} \quad 
    \beta = \begin{cases}
        p \text{ if } Lat(a,h) \leq l_h \\
        q \text{ otherwise}
    \end{cases} 
    \label{eq:reward_beta}
\end{equation}
We use an in-house cycle-accurate performance simulator to estimate the on-device latency of sampled architectures during NAS. 
This offers a middle ground between the accurate-but-expensive benchmarking methods that use real, physical devices and the cheap-but-inaccurate one that use proxy metrics like FLOPs, MACs, or number of parameters. 
Moreover, by configuring such a simulator differently, we can inject hardware-specific information and bias the search to adapt to different targeted edge platforms.

Unlike prior works \cite{mnasnet,nahas}, we extend the notion of $Acc()$ to multi-task setting using a simple-yet-effective nested weighted product of metrics and tasks. 
Let $M_{i} = \{m_{i,1}, m_{i,2}, ..., m_{i,K} \}$ be the set of metrics of interest for tasks $T_i$, e.g. \{mIoU, PixelAcc\} for semantic segmentation. Our multi-task $Acc()$ can be expressed as:
\begin{equation}
  Acc(a, T, w_a) = \left[\prod_i m_i \right]^{1/N}
  \;
  \label{eq:reward_acc_all}
\end{equation}
\begin{equation}
  \text{s.t.} \;\;
  m_i = \left[ \prod_j m_{i,j}^{w_{i,j}} \right] ^ {1 / \sum_j w_{i,j}}
  \label{eq:reward_acc_task}
\end{equation}
This extended formulation is straightforward and scalable even when the number of tasks or metrics increases.
%
Since our goal is to discover multi-task networks that can perform well across all tasks without bias to individual tasks, we treat all task rewards equally in our formulation.

\paragraph{Edge-Friendly Base Architecture}
Previously works~\cite{bmtas,learning2branch,mtan,adashare,branched_mt} typically use bigger networks such as ResNet \cite{resnet} or VGG \cite{vgg} backbone with ASPP \cite{deeplab_v2_2017} decoder. Such models, however, are not suitable for edge platforms like the Coral TPU \cite{coral_edgetpu} due to their limited computational resources.
To this end, we propose the use of EfficientNet~\cite{efficientnetv1,efficientnetv2} backbone and BiFPN fusion modules \cite{efficientdet}, which have been shown to have significantly better FLOPs and parameter efficiency (e.g. an order of magnitude lower) compared to their counterparts \cite{dynamic_routing_seg,efficientnetv1,efficientdet,nahas}.
These advantages make them promising candidate modules to build edge-friendly models.
To generate multi-task outputs while saving computation, we share the majority of the network, including both the EfficientNet backbone and BiFPN modules, across all tasks and use only small per-task heads.
This keeps our model compact and avoids a significant increase in size as the number of tasks goes up .
We also replace Swish activation and attention-based fusion with ReLU6 and Sum operations in \cite{efficientnetv1} to further improve efficiency on edge.
We balance the compact EfficientNet backbone with 4 BiFPN fusion modules instead of 3 like \cite{efficientdet} to boost accuracy. 
The multi-scale fusion modules take features $\{P_3, P_4, P_5, P_6, P_7\}$ from levels 3-7 of the backbone.
These components together make up our edge-friendly base architecture, which we will use as both the seed for our NAS and the baseline model for evaluating MTL performance.

\vspace{1em}
\paragraph{Edge-Friendly Search Space}
Modern NAS usually retains some aspect of the base architecture in order to keep the search space tractable and to reduce the computational cost. Thus, it is important to have a good initialization architecture to seed the search.
For this, 
we leverage the base architecture designed above and
Pyglove \cite{pyglove}, a Python AutoML library that supports flexible layer-level mutation for NAS components via symbolic programming. 
This allows us to transform the static EfficientNet backbone into a tunable search space by replacing any standard computational node with a PyGlove's mutable object, e.g. converting {\tt Conv2d(kernel=3)} into {\tt Conv2d(kernel=oneof([3,5,7]))}.
Furthermore, we expand the search space to include Fused-IBN~\cite{efficientnetv2,mobiledet,nahas} modules 
alongside the standard Inverted Bottleneck (IBN) \cite{mobilenetv2}.
Despite inciting more trainable parameters, Fused-IBN can potentially offer better efficiency on edge devices if strategically placed, e.g. via NAS. This is because industry accelerators are better tuned for regular convolution than their depthwise counterparts, e.g. resulting in 3$\times$ speedup for certain tensor shapes and kernel dimensions~\cite{mobiledet}.
Our final search space is defined by the following per-layer decisions:
\begin{itemize}
\setlength\itemsep{-.3em}
    \item Layer type: \{IBN, Fused-IBN\}
    \item Kernel size: \{3, 5\}
    \item Output channel multiplier: \{0.5, 0.75, 1.0, 1.5\}
    \item Expansion ratio: \{3, 6\}
\end{itemize}
The search is performed for all 16 IBN blocks of our base EfficientNet backbone, together with the other search parameters, producing an expressive search space of size  $(2*2*4*2)^{16} = 2^{80} \approx 1.2e24$.

\vspace{1em}
\subsection{Depth Estimation Noise and JAReD Loss}
\vspace{-0.5em}
\input{tab/depth_mre}
\paragraph{Instability in Depth Estimation}
During our study, we discover that depth prediction accuracy can vary greatly across different training runs of the same setting.
This is illustrated in \Tab{depth_mre} by the results of standard depth training with $ \mathcal{L}_1$ loss.
Note that the standard deviation of depth errors across identical runs are fairly large at 4.4\% and 4.1\%, $\times 2$ higher than that of segmentation mIoU.
%
Such large variation is problematic for the multi-task evaluation as one model could potentially arbitrarily and falsely ``improve'' or ``degrade" purely by chance.
Moreover, this may even interfere with the joint learning MT-DP and NAS through noisy task accuracy in the objective function in Eq \ref{eq:reward}.
In other words, it would be challenging for NAS to identify good architectures if training accuracy itself is unstable and unreliable.
%
%
%

\paragraph{Joint Absolute-Relative Depth}
We hypothesize that the noisy depth result is due to
the fact that popular MT-DP training \cite{mtan,adashare,survey_mt_dense} relies only $\mathcal{L}_1$ loss, which focuses on optimizing for absolute depth and only implicitly learn relative depth.
For monocular setting, learning absolute depth directly is ill-posed and challenging due to the scale ambiguity ~\cite{scale_ambiguity, rel_depth}.
Instead, we propose to augment the standard loss using a weighted relative-error component, resulting in a Joint Absolute-Relative Depth loss, or \textit{JAReD}:
\begin{equation} 
    \mathcal{L}_{JAReD} 
    = \frac{1}{N} \Sigma |y - \hat{y}| + 
    \lambda \frac{1}{N} \Sigma \left|\frac{y - \hat{y}}{y} \right|
    \label{eq:depth_loss}
\end{equation}
\Tab{depth_mre} shows that JAReD can help significantly reduce depth estimation noise---the STDs of all tasks decrease,
especially for relative error with 87.8\% lower fluctuation.
Moreover, JAReD can simultaneously improve accuracy, with both absolute and relative errors dropping by 4.7\% and 8.6\%.

%% file: tab/depth_mre.tex
\begin{table}
\centering
\resizebox{\linewidth}{!}{ 
    \begin{tabular}{@{}l|cccccc@{}}
        \toprule
        Depth Loss
            & mIoU
            & $\sigma\%$
            & AbsE 
            & $\sigma\%$
            & RelE 
            & $\sigma\%$ \\
        \midrule
        $\mathcal{L}_1$
            & 38.6
            & 2.5
            & 0.01763 
            & 4.4
            & 0.3541
            & 4.1
            \\
        JAReD
            & 38.9
            & 1.6
            & 0.01680
            & 1.9
            & 0.3237 
            & 0.5
            \\
        \textit{\% improved}
            & $\uparrow$\textit{0.8}
            & $\downarrow$\textit{36.0}
            & $\downarrow$\textit{4.7}
            & $\downarrow$\textit{56.8}
            & $\downarrow$\textit{8.6}
            & $\downarrow$\textit{87.8}
            \\
        \bottomrule
    \end{tabular}
} 
\caption{
    \textbf{Depth estimation noise and JAReD loss.}
    We train a model for segmentation and depth prediction on CityScapes with the standard  $\mathcal{L}_1$ and proposed JAReD loss. The mean and standard deviation are taken across 3 runs. 
    Except for mIoU, lower is better.
}
\label{tab:depth_mre}
\end{table}


%% file: sec/4_results.tex
\vspace{1em}
\section{Experiments}
\label{sec:experiments}
\input{tab/sota_cityscapes}
\subsection{Setup}
\vspace{-0.5em}
\paragraph{Datasets and Tasks}
We evaluate our proposed method using two popular datasets for multi-task dense predictions: CityScapes \cite{cityscapes} and NYU-v2 \cite{nyuv2}.
CityScapes contains 2975 training images and 500 validation images of driving scenes while NYU-v2 is composed of 1449 densely labeled RGBD indoor images, with a stand training-to-validation split of 795 to 654.
We use the preprocessed versions provided by AdaShare \cite{adashare}.
We jointly learn semantic segmentation (19 classes) and depth prediction for CityScapes.
For NYU-v2, we study 3-task learning of segmentation, depth prediction, and surface normal estimation.

\paragraph{Baselines} 
We adopt the standard practice of evaluating our proposed techniques against the Single-Task (ST) and vanilla Multi-Task (MT) versions, which are EfficientNet-based in our case. We refer to these as \textit{edge baselines}. 
For fair comparisons, we consult the training hyperparameters used by AdaShare \cite{adashare} to match their baseline performance and only compare the relative improvements. 

\paragraph{Implementation Details}
For all experiments, we use EfficientNet-B0 \cite{efficientnetv1} as our backbone. 
We use Regularized Evolution \cite{reg_evol} as our search controller as it can produce compact and accurate models with less search time, thus shortening the experimentation cycle.
Nonetheless, we expect other controllers, e.g. PPO \cite{ppo} as used by prior works~\cite{mnasnet,nahas}, to also work.
%
%
We use Adam \cite{adam} optimizer and cosine learning rate scheduler for all our training, including both the proxy task during NAS and the final training of the best candidates, to reduce hyperparameter tuning effort.
For full training, we train each model 3 times and take the average results similar to \Table{depth_mre} to reduce noise.
All models are trained from scratch without any pretrained weights.
We acquire wall-clock latency measurements by benchmarking models on a Coral EdgeTPU \cite{coral_edgetpu}.
Further details are included in the supplementary.

\paragraph{Evaluation Metrics}
We use mean Intersection over Union (mIoU) and pixel accuracy (PAcc) for semantic segmentation, and mean absolute error (AbsE) and mean relative error (RelE) for depth prediction. 
For surface normal estimation on NYU-v2, we use mean angle distance error (MeanE) across all pixels, as well as the percentage of pixels with angle distances less a threshold $\theta \in \{11.25\degree , 22.5\degree , 30\degree\}$, denoted as \{$\theta$11, $\theta$22, $\theta$30\} respectively.
Following other works \cite{single_tasking_attentive,adashare,survey_mt_dense}, we calculate a single evaluation score $\Delta T$ averaging over all relative gains $\Delta T_i$ of all tasks $T_i$ relative to the Single-Task baseline.
A formal definition of these metrics are provided in our supplementary materials. 
\input{tab/x_nyu_sdn_baselines2}
\input{tab/x_nyu_segdepthnorm}
\input{tab/2_gen_stronger_cityscapes} 

\vspace{1em}
\subsection{Results}
\vspace{-0.5em}
\paragraph{EDNAS for 2-task CityScapes}
\Tab{main_table} shows our experiments for the 2-task learning of 19-class semantic segmentation and depth estimation on CityScapes dataset. 
In this experiment,
the same $\Delta T$ of -4.1 is shared by the MT edge baseline and its large-scale counterpart, indicating that they both experience a similar level of negative transfer and MTL difficulty.
Following \cite{adashare}, we present MTL gains relative to the ST baseline model.
The proposed EDNAS exhibits a strong multi-task performance with $\Delta T$=+8.5, outperforming all prior methods. 
Since the full training of MT edge baseline and EDNAS-found architecture are \textit{identical}, it shows that joint MTL-DP and NAS can produce a superior relative improvement of +8.5 - (-4.1) = +12.6 compared to the vanilla multi-task model. 

\paragraph{JAReD Loss}
From \Tab{main_table}, we see that the proposed JAReD loss is able to greatly improve depth estimation with a relative gain of $\Delta T_D$=13.3\%. This in turn further strengthens the overall multi-task performance by a significant margin of +2.4 on top of the already-strong result ($\Delta T$=+8.5) of EDNAS. 
Together, our two proposed techniques outperform all previous approaches on 3 out of 4 individual metrics, namely $\Delta$mIoU, $\Delta$PAcc, and $\Delta$RelE, as well as on all the average metrics, which are $\Delta T_S$, $\Delta T_D$, and $\Delta T$.

\paragraph{Edge-Efficient Inference} 
Regarding edge efficiency, EDNAS and EDNAS+JAReD use only 1/5th of the parameters and 1/10th of the FLOPs compared to prior ResNet-based methods. More importantly, the EDNAS-found model is able to practically maintain the same on-device speed as the vanilla MT baseline, if not slightly faster, despite the +12.6 improvement. This equates to a 30\% improvement in latency compared to separate single-task inferences, and further demonstrates the benefits of our proposed joint learning for discovering and training better multi-task architectures for dense predictions on edge platforms.

\paragraph{Generalization to 3-Task NYUv2}
%
Unlike with Cityscapes where the MT baselines have similar accuracy drop,
for NYUv2, we notice a large difference between the amount of negative transfer in MT edge baseline ($\Delta T$=-11.3) and in the large-scale MT model ($\Delta T$=+2.0), as shown in \Tab{nyu_sdn_baselines}.
This indicates that multi-task training on NYUv2 data may be more challenging for edge models with limited computation. 
Because of such discrepancy in the level of MTL difficulty, we directly use the MT models (instead of ST models) as the baselines to benchmark the improvement gained.
Note that despite such a large gap compared to the ST edge setting, our MT edge model is still comparable to the computationally heavy ST baselines of prior studies, with a negligible {$\Delta T$=-0.1}.
%
The NYUv2 results from \Tab{nyu_segdepthnorm} show that EDNAS and JAReD continue to achieve consistent and significant improvements ($\Delta T$ of +9.6 and +12.7) over the baseline, similar to what we observed for Cityscapes.

\paragraph{Robustness to Stronger Baselines}
To further demonstrate the robustness of EDNAS as a solution for discovering better multi-task architectures for dense predictions, we are interested in examining its performance with stronger baselines (\Tab{2_gen_stronger_cityscapes}).
Although prior work \cite{adashare} only uses learning rates in the order of 1e-4 to 1e-3, we also experiment with other rates and observe a huge jump of $\Delta T$=+20.3 in performance when simply increasing the learning rate while holding other settings the same.
We utilize this simple adjustment to obtain our stronger edge baseline with the largest learning rate of maxLR=1e-2.
Taking a step further, we add JAReD loss to our ST edge baseline both to demonstrate the effectiveness of JAReD loss even for single-task depth estimation and to acquire our strongest baseline for evaluation.
Our result of training the EDNAS-found architecture with similar setup (+maxLR and +JAReD) illustrates the strength of our proposed method with a relative multi-task gain of $\Delta T$=+3.3.
We emphasize that +3.3, despite being smaller than the improvements we have seen so far, is \textit{still} comparable to the majority of state-of-the-art methods shown in \Tab{main_table}, and that is on top of a \textit{+30\% stronger} ST baseline!
 

\paragraph{Joint Learning vs Transfer Learning}
\Tab{2_gen_stronger_cityscapes} also shows the performance of EDNAS when compared to the transferring of NAS-found single-task models to the multi-task setting. 
Although transferred architectures can bring a considerable amount of improvement compared to our baseline ST and MT models, EDNAS' joint learning of multi-task dense predictions and hardware-aware NAS evidently offers the optimal performance among these models, achieving either the best or second best scores in all categories.
Moreover, it is also important to note that there is a significant difference in the performance gains 
of the transferred depth estimation network compared to that of the transferred segmentation model.
Therefore, we may not know in advance which specific tasks transfer better than the other, further illustrating the power and benefits of our EDNAS.



\paragraph{Analysis of EDNAS-Found Architectures}
\input{tab/x_arch_found_segdepth}
\input{tab/x_arch_found_depth}
\Tab{arch_found}~~gives a summary of the backbone architecture found by EDNAS for multi-task segmentation and depth estimation on CityScapes. 
This is the same model as presented in CityScapes experiment section.
Except for the first Conv2D layer, which is a fixed stem, the following 16 layers (1-16) are all tunable.
Our first observation is that FusedIBN is heavily favored by the search algorithm over regular IBN, occupying 14 out of 16 tunable layers. This is likely due to the fact that modern edge accelerators such as the Coral Edge TPU \cite{coral_edgetpu} are more optimized for normal convolution than for depthwise separable convolution. Therefore, they can leverage the dense computations to improve both accuracy and inference latency.
Second, we notice that 4 out of our top 5 searched models have an IBN module at layer 2 and 7, including the one in \Tab{arch_found}. The remaining architecture also has IBN for layer 7 but not for layer 2. Hence, we believe that even though sparsely used, IBN layers can still be beneficial if placed strategically, e.g. via EDNAS.

\Tab{arch_found_depth} provides an example of architectures found by our single-task NAS for depth estimation.
We observe that there are consistently and considerably lower numbers of FusedIBN modules, namely 11 compared to 14 in Table \Tab{arch_found}, which is produced by EDNAS, a multi-task NAS algorithm.
Similar observation also applies to the single-task NAS for segmentation, which has 12 FusedIBN layers.
We conjecture that multi-task learning might require more powerful and expressive layers to capture cross-task nuances.
As a result, single-task NAS, which performs an indirect search using individual tasks, may fail to recognize and meet these needs, leading to fewer FusedIBN blocks and poorer accuracy as seen in the transferring experiments.



%% file: tab/sota_cityscapes.tex
\begin{table*}[!t]
\centering
\resizebox{\linewidth}{!}{ 
\begin{tabular}{@{}lccc|cccc|cccc|ccc@{}}

\toprule

\multicolumn{1}{c}{}  
    & \multicolumn{3}{c}{Model} 
    & \multicolumn{2}{c}{Seg} 
    & \multicolumn{2}{c}{Depth} 
    & \multicolumn{2}{c}{$\Delta$Seg} 
    & \multicolumn{2}{c}{$\Delta$Depth} 
    & \multicolumn{3}{c}{Avg} \\

Method 
    & \#P
    & GFLOP
    & Speed
    & mIoU 
    & PAcc 
    & AbsE 
    & RelE 
    & $\Delta$mIoU 
    & $\Delta$PAcc 
    & $\Delta$AbsE 
    & $\Delta$RelE
    & $\Delta T_{S}$
    & $\Delta T_{D}$
    & $\Delta T$
    \\

\midrule

ST baseline \cite{adashare}  
    & 42.6
    & 87.1 
    & ---
    & 40.20  
    & 74.70  
    & .0170 
    & .330 
    & --- & --- & --- & ---
    & ---
    & ---
    & ---
    \\
MT baseline \cite{adashare} 
    & 21.3
    & 43.6
    & ---
    & 37.70 
    & 73.80 
    & .0180 
    & .340 
    & -6.2 & -1.2 & -5.9 & -3.0
    & -3.7 
    & -4.5 
    & -4.1 
    \\

Cross-Stitch \cite{cross_stitch} 
    & 42.6
    & 48.4
    & ---
    & 40.30 
    & 74.30 
    & .0150 
    & .300 
    & +0.2 
    & -0.5 
    & \textbf{+11.8} 
    & \underline{+9.1}
    & -0.1 
    & \underline{+10.4} 
    & {+5.1} 
    \\

Sluice \cite{sluice} 
    & 42.6
    & 48.4
    & ---
    & 39.80 
    & 74.20 
    & .0160 
    & .310 
    & -1.0 & -0.7 & 5.9 & 6.1
    & -0.8 
    & +6.0 
    & +2.6 
    \\

NDDR-CNN \cite{nddr} 
    & 44.1
    & 50.1 
    & ---
    & 41.50 
    & 74.20 
    & .0170 
    & .310 
    & 3.2 & -0.7 & 0.0 & 6.1
    & +1.3 
    & +3.0 
    & +2.2 
    \\

MTAN \cite{mtan} 
    & 51.3
    & 57.9 
    & ---
    & 40.80 
    & 74.30 
    & .0150 
    & .320 
    & +1.5 & -0.5 & \textbf{+11.8} & +3.0
    & +0.5 
    & {+7.4} 
    & {+3.9} 
    \\
    
DEN \cite{nas_mt_cls_elastic} 
    & 23.9
    & 51.2 
    & ---
    & 38.00 
    & 74.20 
    & .0170 
    & .370 
    & -5.5 & -0.7 & 0.0 & -12.1
    & -3.1 
    & -6.1 
    & -4.6 
    \\

AdaShare \cite{adashare} 
    & 21.3
    & 87.1 
    & ---
    & 41.50 
    & 74.90 
    & .0160 
    & .330 
    & 3.2 & 0.3 & 5.9 & 0.0
    & {+1.8} 
    & +2.9 
    & +2.3 
    \\

\midrule

ST edge baseline 
    & 3.4 
    & 2.3 
    & $\times$1.0
    & 40.04	
    & 88.68
    & .0157  
    & .340 
    & ---&---&---&---
    & ---
    & ---
    & ---
    \\
MT edge baseline 
    & 3.4 
    & 1.2 
    & $\times$1.2
    & 38.64	 
    & 88.49	
    & .0171	
    & .354 
    & -3.5 & -0.2 & -8.5 & -4.1
    & -1.9	
    & -6.3	
    & -4.1	
    \\
EDNAS 
    & 4.3 
    & 4.1 
    & $\times$1.3
    & 46.52	
    & 90.61	
    & .0143	
    & .316
    & +\textbf{16.2} 
    & +\textbf{2.2} 
    & +8.9 
    & +6.9
    & +\textbf{9.2}	
    & +7.9	
    & +\underline{8.5}	
    \\ 
EDNAS+JAReD 
    & 4.3 
    & 4.1 
    & $\times$1.3
    & 46.11	
    & 90.47		
    & {.0143}
    & {.281}
    & +\underline{15.1} 
    & +\underline{2.0} 
    & +\underline{9.1} 
    & +\textbf{17.4}
    & +\underline{8.6}			
    & +\textbf{13.3}	
    & +\textbf{10.9}		
    \\

\bottomrule
\end{tabular}
} 

\caption{
\textbf{Two-task CityScapes results.}
Best numbers are in \textbf{bold}, the second best are \underline{underlined}. 
\textit{ST} stands for single-tasks. \textit{MT} stands for multi-task. We multiply the FLOPs by the number of tasks for methods that need multiple runs to get different per-task predictions. FLOP counts are in gigas(G) and parameter counts are in millions(M). Both of these, along with our model's edge latency, are measured for 256x256 resolution. We consult Table 8 and Table 11 in \cite{adashare} as well as its first author to acquire the full measurements of prior works
} 
\label{tab:main_table}
\end{table*}

%% file: tab/x_nyu_sdn_baselines2.tex
\begin{table}[t]
\centering

\resizebox{\linewidth}{!}{ 
\begin{tabular}{@{}lcccccccc|c@{}}
\toprule
\multicolumn{1}{c}{}  
    & \multicolumn{2}{c}{Seg} 
    & \multicolumn{2}{c}{Depth} 
    & \multicolumn{4}{c}{Surface Normal} 
    & \multicolumn{1}{c}{Avg} 
    \\
Method 
    & mIoU 
    & PAcc 
    & AbsE 
    & RelE 
    & MeanE 
    & $\theta$11 
    & $\theta$22 
    & $\theta$30  
    & $\Delta T$
    \\

\midrule

ST edge
	& 23.1
	& 58.3
	& 0.50
	& 0.20
	& 13.8
	& 50.8
	& 81.2
	& 90.8
	& ---
	\\
MT edge
	& 19.5
	& 54.8
	& 0.55
	& 0.22
	& 16.5
	& 41.9
	& 73.0
	& 85.1
	& {-11.3}
	\\
\midrule

ST \cite{adashare}
	& 27.5
	& 58.9
	& 0.62
	& 0.25
	& 17.5
	& 34.9
	& 73.3
	& 85.7
	& ---
	\\
MT \cite{adashare}
	& 24.1
	& 57.2
	& 0.58
	& 0.23
	& 16.6
	& 42.5
	& 73.2
	& 84.6
	& {+2.0}
	\\
MT edge
	& 19.5
	& 54.8
	& 0.55
	& 0.22
	& 16.5
	& 41.9
	& 73.0
	& 85.1
	& {-0.1}
	\\
\bottomrule
\end{tabular}
} 

\caption{
\textbf{NYUv2 baselines.}
\textit{ST} and \textit{MT} are prior large-scale models from \cite{adashare} while \textit{edge} denotes our edge-friendly baselines
} 
\label{tab:nyu_sdn_baselines}
\vspace{-1em}
\end{table}

%% file: tab/x_nyu_segdepthnorm.tex
\begin{table*}[t]
\centering

\resizebox{\linewidth}{!}{ 
\begin{tabular}{@{}lcccccccccccc@{}}
\toprule
\multicolumn{1}{c}{}  
    & \multicolumn{2}{c}{Seg} 
    & \multicolumn{2}{c}{Depth} 
    & \multicolumn{4}{c}{Surface Normal} 
    & \multicolumn{4}{c}{Avg} 
    \\
Method 
    & mIoU 
    & PAcc 
    & AbsE 
    & RelE 
    & MeanE 
    & $\theta$11 
    & $\theta$22 
    & $\theta$30  
    & $\Delta T_{S}$
    & $\Delta T_{D}$
    & $\Delta T_{SN}$
    & $\Delta T$
    \\

\midrule

MT baseline \cite{adashare}
	& 24.1
	& 57.2
	& 0.58
	& 0.23
	& 16.6
	& 42.5
	& 73.2
	& 84.6
	& ---
	& ---
	& ---
	& ---
	\\

Cross-Stitch \cite{cross_stitch}
    & 25.4 & 57.6 & 0.58 & 0.23 & 17.2 & 41.4 & 70.5 & 82.9 
	&+3.0
	&+0.0
	&-3.0
	&+0.0 
    \\
Sluice \cite{sluice} 
    & 23.8 & 56.9 & 0.58 & 0.24 & 17.2 & 38.9 & 71.8 & 83.9 
	&-0.9
	&-2.2
	&-3.7
	&-2.3
    \\
NDDR-CNN \cite{nddr} 
    & 21.6 & 53.9 & 0.66 & 0.26 & 17.1 & 37.4 & 73.7 & 85.6 
	&-8.1
	&-13.4
	&-3.3
	&-8.3
    \\
MTAN \cite{mtan}
    & 26.0 & 57.2 & 0.57 & 0.25 & 16.6 & 43.7 & 73.3 & 84.4 
	&+3.9
	&-3.5
	&+0.7
	&+0.4
    \\
DEN \cite{nas_mt_cls_elastic} 
    & 23.9 & 54.9 & 0.97 & 0.31 & 17.1 & 36.0 & 73.4 & 85.9 
	&-2.4
	&-51.0
	&-4.1
	&-19.2
    \\

AdaShare \cite{adashare}
	& 30.2
	& 62.4
	& 0.55
	& 0.20
	& 16.6
	& 45.0
	& 71.7
	& 83.0
	&\textbf{+17.2}
	&\textbf{+9.1}
	&+0.5
	&+8.9
	\\

\midrule

MT edge baseline
	& 19.5
	& 54.8
	& 0.55
	& 0.22
	& 16.5
	& 41.9
	& 73.0
	& 85.1
	& ---
	& ---
	& ---
	& ---
	\\
EDNAS 
	& 22.1
	& 57.7
	& 0.51
	& 0.20
	& 14.3
	& 49.5
	& 79.2
	& 89.4
	&+9.3
	&\underline{+8.2}
	&\underline{+11.3}
	&\underline{+9.6}
	\\
EDNAS+JAReD 
	& 22.1
	& 58.1
	& 0.51
	& 0.20
	& 12.6
	& 56.1
	& 83.9
	& 92.4
	&\underline{+9.7}
	&\underline{+8.2}
	&\textbf{+20.3}
	&\textbf{+12.7}
	\\
\bottomrule
\end{tabular}
} 

\caption{
\textbf{Three-task NYUv2 results 
.}
Our tasks of interest include 40-class semantic segmentation, depth estimation, and surface normal estimation. Best numbers are in \textbf{bold}, the second best are \underline{underlined}.
\textit{ST} stands for single-task and \textit{MT} stands for multi-task.
We multiply the FLOPs by the number of tasks for methods that need multiple runs to get different per-task predictions.
The measurements of prior works are from Table 9 and Table 11 in \cite{adashare}
} 
\label{tab:nyu_segdepthnorm}

\end{table*}

%% file: tab/2_gen_stronger_cityscapes.tex
\begin{table*}[!t]
\centering

\resizebox{\linewidth}{!}{ 
\begin{tabular}{@{}lccc|cccc|cccc|ccc@{}}

\toprule

\multicolumn{1}{c}{}  
    & \multicolumn{3}{c}{Model} 
    & \multicolumn{2}{c}{Seg} 
    & \multicolumn{2}{c}{Depth} 
    & \multicolumn{2}{c}{Seg$\Delta$} 
    & \multicolumn{2}{c}{Depth$\Delta$} 
    & \multicolumn{3}{c}{Avg$\Delta$} \\

Method 
    & \#P
    & GFLOP
    & Speed
    & mIoU 
    & PAcc 
    & AbsE 
    & RelE 
    & $\Delta$mIoU 
    & $\Delta$PAcc 
    & $\Delta$AbsE 
    & $\Delta$RelE
    & $\Delta T_{S}$
    & $\Delta T_{D}$
    & $\Delta T$
    \\

\midrule


ST edge baseline 
    & 3.4 
    & 2.3 
    & $\times$1.0
    & 40.04	
    & 88.68
    & .0157  
    & .340 
    & ---&---&---&---
    & ---
    & ---
    & ---
    \\
    
ST edge+maxLR 
    & 3.4 
    & 2.3 
    & $\times$1.0
    & {55.02} 
    & {92.29} 
    & .0121
    & .288
    & +37.4	
    & +4.1	
    & +23.2	
    & +15.3	
    & +20.7	
    & +19.3	
    & +20.0
    \\
    
ST edge+maxLR+JAReD 
    & 3.4 
    & 2.3 
    & $\times$1.0
    & {55.02} 
    & {92.29} 
    & \textbf{.0116}   
    & .168 
    & +37.4	
    & +4.1	
    & +26.7	
    & +50.5	
    & +20.7	
    & +38.6	
    & +29.7
    \\



\midrule

ST edge+maxLR+JAReD 
    & 3.4 
    & 2.3 
    & $\times$1.0
    & {55.02} 
    & {92.29} 
    & \textbf{.0116}   
    & .168 
    & ---&---&---&---
    & ---
    & ---
    & ---
    \\
MT edge+maxLR+JAReD 
    & 3.4 
    & 1.2 
    & $\times$1.2
    & 53.80	
    & 91.94	
    & {.0124} 
    & {.159} 
    & -2.2
    & -0.4
    & -7.4	
    & +5.7
    & -1.3   
    & -0.9  
    & -1.1 
    \\
Transfer: NAS-Seg$\xrightarrow{}$MT 
    & 4.1 
    & 2.5 
    & $\times$1.3
    & \underline{58.17}
    & \textbf{92.78}
    & .0118
    & \textbf{.156}
    & +5.4	
    & \underline{+0.5}	
    & -2.9	
    & \underline{+6.2}
    & \underline{+3.1} 
    & \underline{+2.8} 
    & \underline{+3.0} 
    \\
Transfer: NAS-Dep$\xrightarrow{}$MT 
    & 3.6 
    & 2.5 
    & $\times$1.3
    & 57.97
    & \underline{92.73}
    & .0119
    & \underline{.158} 
    & \underline{+5.7}
    & \underline{+0.5}	
    & \underline{-1.8}	
    & \textbf{+7.4}
    & +2.9
    & +1.6 	
    & +2.3
    \\
EDNAS+maxLR+JAReD
    & 4.3 
    & 4.1 
    & $\times$1.3
    & \textbf{58.54} 
    & \textbf{92.78} 
    & \underline{.0117}   
    & \textbf{.156} 
    & \textbf{+6.4}	
    & \textbf{+0.5}	
    & \textbf{-1.3}	
    & \textbf{+7.4}
    & \textbf{+3.5}
    & \textbf{+3.1}  
    & \textbf{+3.3}
    \\ 

\bottomrule
\end{tabular}
} 

\caption{
\textbf{Stronger baselines on CityScapes.} 
\textit{ST edge baseline} and \textit{ST edge+maxLR} have identical training setting with the only exception of their learning rate being 3e-4 and 1e-2 respectively
} 
\label{tab:2_gen_stronger_cityscapes}

\end{table*}

%% file: tab/x_arch_found_segdepth.tex
\begin{table}
\centering
\resizebox{0.93\linewidth}{!}{ 
\begin{tabular}{@{}cccccc@{}}
\toprule
Index & Layer & Stride & Kernel & Filters & Expansion \\
\midrule
0 & Conv2D & 2 & 3 & 32  & -- \\
1 & FusedIBN & 1 & 3 & 16  & 1 \\
2 & IBN & 2 & 5 & 36  & 6 \\
3 & FusedIBN & 1 & 5 & 24  & 6 \\
4 & FusedIBN & 2 & 3 & 60  & 6 \\
5 & FusedIBN & 1 & 3 & 40  & 3 \\
6 & FusedIBN & 2 & 5 & 120 & 3 \\
7 & IBN & 1 & 3 & 120 & 3 \\
8 & FusedIBN & 1 & 5 & 80  & 6 \\
9 & FusedIBN & 1 & 5 & 168 & 6 \\
10 & FusedIBN & 1 & 5 & 84  & 3 \\
11 & FusedIBN & 1 & 5 & 84  & 6 \\
12 & FusedIBN & 2 & 5 & 288 & 3 \\
13 & FusedIBN & 1 & 3 & 96  & 3 \\
14 & FusedIBN & 1 & 3 & 96  & 6 \\
15 & FusedIBN & 1 & 3 & 96  & 3 \\
16 & FusedIBN & 1 & 5 & 160 & 6 \\
\bottomrule
\end{tabular}
} 
\caption{
\textbf{Backbone Architecture found by EDNAS} -- 
Backbone architecture found EDNAS for multi-task segmentation and depth estimation on CityScapes, same model as presented in \Tab{main_table}.
} 
\label{tab:arch_found}
\vspace{-1em}
\end{table}

%% file: tab/x_arch_found_depth.tex
\begin{table}
\centering
\resizebox{0.93\linewidth}{!}{ 
\begin{tabular}{@{}cccccc@{}}
\toprule
Index & Layer & Stride & Kernel & Filters & Expansion \\
\midrule
0 & Conv2D & 2 & 3 & 32  & -- \\
1 & FusedIBN & 1 & 3 & 24  & 1 \\
2 & IBN & 2 & 3 & 36  & 6 \\
3 & IBN & 1 & 3 & 36  & 6 \\
4 & FusedIBN & 2 & 5 & 40  & 6 \\
5 & FusedIBN & 1 & 5 & 40  & 3 \\
6 & IBN & 2 & 3 & 80  & 6 \\
7 & FusedIBN & 1 & 3 & 120 & 3 \\
8 & FusedIBN & 1 & 3 & 80  & 6 \\
9 & FusedIBN & 1 & 3 & 168 & 3 \\
10 & FusedIBN & 1 & 3 & 56  & 6 \\
11 & FusedIBN & 1 & 3 & 112 & 3 \\
12 & FusedIBN & 2 & 5 & 192 & 6 \\
13 & FusedIBN & 1 & 3 & 192 & 6 \\
14 & IBN & 1 & 5 & 192 & 3 \\
15 & IBN & 1 & 5 & 192 & 3 \\
16 & FusedIBN & 1 & 5 & 240 & 3 \\
\bottomrule
\end{tabular}
} 
\caption{
\textbf{Backbone Architecture found by Single-task NAS} -- 
An example of the backbone architecture found the single-task NAS targeting depth estimation on CityScapes 
. We suspect that multi-task learning can benefit from more expressive layers such as FusedIBN; thus, fewer of such layers compared \Tab{arch_found} may correlate to the lower accuracy 
as seen in the previous experiments.
} 
\label{tab:arch_found_depth}
\vspace{-1em}
\end{table}

%% file: sec/5_conclusions.tex
\section{Conclusion}
In this work, our two main contributions include EDNAS and JAReD loss. The former is a novel and scalable solution that exploits the synergy of MTL and h-NAS to improve both accuracy and speed for dense prediction task on edge platforms. 
The latter is an easy-to-adopt augmented depth loss that simultaneously mitigates noise and further boosts accuracy.
Through extensive experimentation, we show that the proposed techniques can outperform state-of-the-art methods, minimize on-device computational cost, generalize to different data and training settings, as well as discover meaningful and effective architectures.

%% file: sec/X_supplementary.tex
\appendix

\twocolumn[
    \centering
    \Large
    \textbf{Toward Edge-Efficient Dense Predictions
    with Synergistic Multi-Task Neural Architecture Search} \\
    \vspace{1em}
    Supplementary Material \\
    \vspace{1.5em}
] 
\appendix


\input{fig/system}

\section{Experimental details}
\paragraph{Hyperparameters of NAS} 
We use a Regularized Evolution
controller with a population size of 50, random initialization, uniform mutator, and a tournament sample size of 10. 
We let the search run for about 2000 generations.
These parameters were simply chosen to fit our computational budget and were not tuned.
During the search, we train models for 5000 iterations as a proxy task to save computation.
The final models are trained for 20000 iterations following AdaShare.
%
%
For the $\beta$ in the objective function in \Eq{reward_beta}, we use (p=0.0) to set up a hard constraint function and (q=-0.07) to promote Pareto optimality, following MnasNet.
We use $w_{i,j}$=1.0 to equally weight all evaluation metrics $M_{i,j}$ of any task $T_i$ in \Eq{reward_acc_all} and \Eq{reward_acc_task}. 
These can be adjusted to suit downstream applications.
%
With 512 TPUv2 cores, our multi-trial search takes about 1.5 days for Cityscapes and 3.5 days for NYUv2.
Since EDNAS is not constrained by the specific NAS algorithm, one can also use a one-shot search with weight sharing \cite{proxylessnas,fbnet} instead for better computational efficiency.
Finally, \Fig{ibn_v_fused} provides a visual comparison of IBN and Fused-IBN blocks.

\input{tab/final_loss_weights} 
\input{tab/loss_weight_impact} 

\paragraph{Task Loss and Weighting}
Following AdaShare \cite{adashare}, we use Cross-Entropy loss $\mathcal{L}_{CE}$ to train semantic segmentation, $\mathcal{L}_1$ loss for the base training of monocular depth estimation, and the inverse of cosine similarity loss $\mathcal{L}_{ICS}$ for surface normal prediction. Our JAReD loss also includes a weighted mean relative error component $\mathcal{L}_{RE}$ as specified in \Eq{depth_loss}
We manually tune the loss weights to avoid ineffective weighting interfering with the evaluation of NAS-found architectures, using two guidelines:
(1) We set task weights so that our \textit{MT edge baseline} best matches AdaShare's (Sec. 4.1), then use similar weights for \textit{EDNAS}.
(2) For \textit{EDNAS+JAReD}, we keep the $\lambda$ in \Eq{depth_loss} small 
to avoid overwhelming the $\mathcal{L}_1$ and other tasks such as segmentation.
\Tab{final_loss_weights} details the final weights of our main models, as presented in \Tab{main_table} and \Tab{nyu_sdn_baselines}.
In addition, \Tab{loss_weight_impact} illustrates the impact of different loss weighting strategies on the multi-task performance of segmentation and depth prediction.


%

\paragraph{$\Delta$ Metrics for MTL Evaluation}
Following the standard metrics for evaluating multi-task learning \cite{single_tasking_attentive,adashare,survey_mt_dense}, we calculate the scores of multi-task learning relative to the single-task performance. 
Specifically, 
given a \textit{multi-task} model $a$ for evaluation,
let $T_i \in T$ be a task of interest (e.g. semantic segmentation) and $m_{ij} \in M_i$ be an evaluation metric for task $T_i$ (e.g. mIoU).
Let $\hat{m}_{ij}$ be the \textit{baseline} score of a corresponding \textit{singe-task} model (e.g. single-task segmentation mIoU).
We define the per-metric relative score $\Delta m_{ij}$ (e.g. $\Delta$mIoU) of the multi-task model $a$ with regard to its baseline $\hat{m}_{ij}$ as followed:
\begin{equation}
    \Delta m_{ij} = (-1)^{l_j} \frac{m_{ij} - \hat{m}_{ij}}{\hat{m}_{ij}} * 100\%
\end{equation}
\begin{equation}
    \text{with }
    l_j = \begin{cases}
      1 \text{ if lower is better for metric $M_j$} \\
      0 \text{ otherwise}
    \end{cases} 
\end{equation}
We then define the per-task relative score $\Delta {T_i}$ (e.g. $\Delta$Seg) of any task $T_i$ and the overall multi-task score $\Delta {T}$ of model $a$ respectively as:
\begin{equation}
    \Delta {T_i} = \frac{1}{|M_i|} \sum_{j=1}^{|M_i|} \Delta m_{ij}
\end{equation}
\begin{equation}
    \Delta T = \frac{1}{|T|} \sum_{i=1}^{|T|} \Delta T_i
\end{equation}
with $|M_i|$ and $|T|$ being the cardinality of the corresponding metric set and task set respectively.



\section{Qualitative Results}
Figure \ref{fig:qualitative_cityscapes} presents some qualitative results of semantic segmentation and depth estimation on CityScapes dataset

\input{fig/ibn_v_fused}

\input{fig/qualitative_cityscapes}



%% file: fig/system.tex
\begin{strip} 
\vspace{-3.5em}
\begin{center}
\captionsetup{type=figure}
\includegraphics[width=\linewidth]{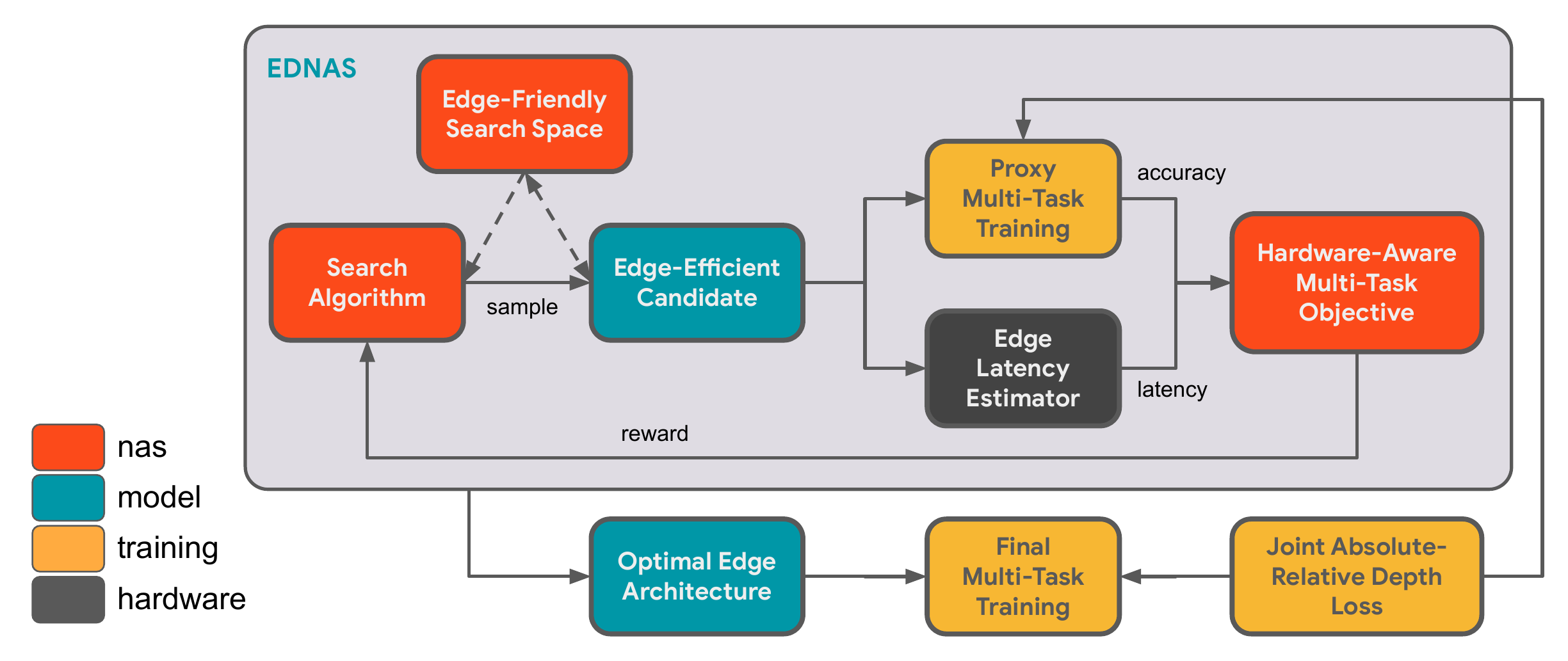}
\captionof{figure}{
    \textbf{A system-level overview of our proposed methods}.
    We leverage multi-objective, hardware-aware neural architecture search to discover optimal neural components suitable for multi-task dense predictions, while simultaneously ensuring efficient edge inference. 
}
\label{fig:system}
\end{center}
\vspace{0.5em}
\end{strip}

%% file: tab/final_loss_weights.tex
\begin{table}
    \centering
    \resizebox{\linewidth}{!}{ 
    \begin{tabular}{@{}l|lcccc@{}}
    \toprule
    \multicolumn{1}{c}{}  
        &\multicolumn{1}{c}{} 
        & \multicolumn{1}{c}{Seg} 
        & \multicolumn{2}{c}{Depth} 
        & \multicolumn{1}{c}{SN}
        \\
    Reference
        & Model 
        & $\mathcal{L}_{CE}$
        & $\mathcal{L}_1$
        & $\mathcal{L}_{RE}$
        & $\mathcal{L}_{ICS}$
        \\
    \midrule
    \Tab{main_table}:
        & MT edge baseline
        & 0.4
        & 1.000
        & ---
        & ---
        \\
    Cityscapes
        & EDNAS
        & 0.4
        & 1.000
        & ---
        & ---
        \\
    ~
        & EDNAS+JAReD
        & 0.5
        & 0.950
        & 0.050
        & ---
        \\
    \midrule
    \Tab{nyu_sdn_baselines}:
        & MT edge baseline
        & 1.0
        & 1.000
        & ---
        & 40.0
        \\
    NYUv2
        & EDNAS
        & 1.0
        & 1.000
        & ---
        & 50.0
        \\
    ~
        & EDNAS+JAReD
        & 1.0
        & 0.999
        & 0.001
        & 60.0
        \\
    \bottomrule
    \end{tabular}
    } 
    \caption{
        \textbf{Final loss weights}. 
        This table specifies the per-task loss weights for models trained on 2-task Cityscapes and 3-task NYUv2. ``SN" stands for surface normal estimation.
    } 
    \label{tab:final_loss_weights}
\end{table}

%% file: tab/loss_weight_impact.tex
\begin{table*}[t]
    \centering
    \resizebox{\linewidth}{!}{ 
    \begin{tabular}{@{}l|ccc|cccc|cc|c@{}}
    \toprule
    \multicolumn{1}{c}{}  
        &\multicolumn{3}{c}{Loss weight} 
        & \multicolumn{2}{c}{Seg} 
        & \multicolumn{2}{c}{Depth} 
        & \multicolumn{3}{c}{Avg} 
        \\
    Method
        & $\mathcal{L}_{CE}$
        & $\mathcal{L}_{1}$
        & $\mathcal{L}_{RE}$
        & mIoU 
        & PAcc 
        & AbsE 
        & RelE
        & $\Delta T_{S}$
        & $\Delta T_{D}$
        & $\Delta T$
        \\
    \midrule
    Single-task seg
        & 1.000
        & 0.000
        & 0.000
        & 40.04 
        & 88.68
        & ---  
        & --- 
        & --- 
        & ---
        & ---
        \\
    Single-task depth
        & 0.000 & 1.000 & 0.000
        & ---
        & ---
        & 0.0157  
        & 0.340 
        & --- 
        & ---
        & ---
        \\
    Multi-task seg-depth
        & 0.500 & 1.000 & 0.000
        & 38.64
        & 88.49 
        & 0.0171 
        & 0.354
        & -1.9 
        & -6.3 
        & -4.1
        \\
    Multi-task seg-depth
        & 0.500 & 0.999 & 0.001
        & 46.78
        & 90.62
        & 0.0149
        & 0.323
        & +9.5
        & +5.1
        & +7.3
        \\
    Multi-task seg-depth
        & 0.500 & 0.990 & 0.010
        & 46.83
        & 90.56	
        & 0.0144	
        & 0.304
        & +9.5
        & +9.5
        & +9.5
        \\
    Multi-task seg-depth
        & 0.500 & 0.950 & 0.050
        & 46.11	
        & 90.47	
        & 0.0143	
        & 0.281
        & +8.6
        & +13.3
        & +10.9
        \\
    Multi-task seg-depth
        & 0.500 & 0.900 & 0.010
        & 46.41	
        & 90.56	
        & 0.0146	
        & 0.300
        & +9.0
        & +9.5
        & +9.3
        \\
    \bottomrule
    \end{tabular}
    } 
    \caption{
        \textbf{Impact of loss weighting}
    } 
    \label{tab:loss_weight_impact}
\end{table*}

%% file: fig/ibn_v_fused.tex
\newcommand{\imgsizeB}{0.48}

\begin{figure}
\begin{center}
    \begin{subfigure}[b]{\imgsizeB\linewidth}
        \centering
        \captionsetup{justification=centering}
        \includegraphics[width=\linewidth]{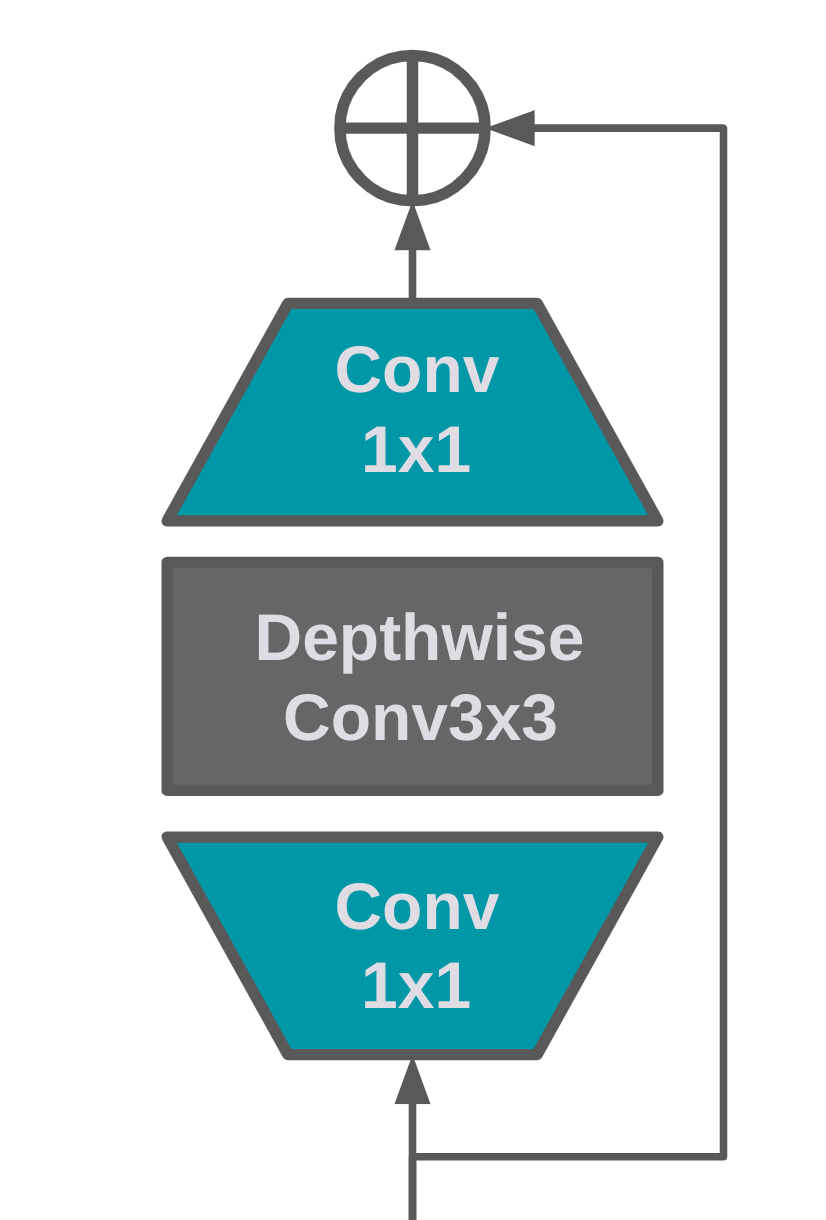}
        \caption*{{(a) Inverted Bottleneck (IBN)}}
    \end{subfigure}
    \\
    \begin{subfigure}[b]{\imgsizeB\linewidth}
        \centering
        \captionsetup{justification=centering}
        \includegraphics[width=\linewidth]{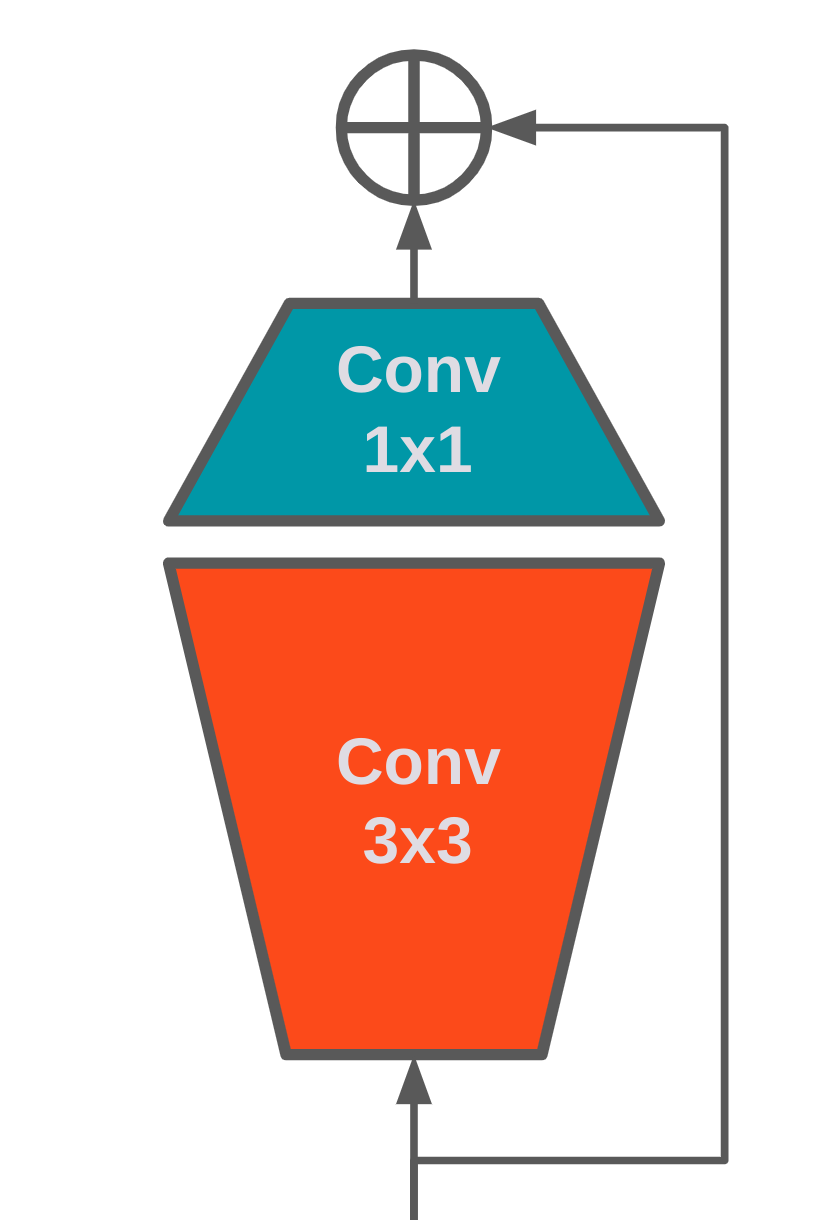}
        \caption*{{(b) Fused-IBN}}
    \end{subfigure}
\end{center}
\caption{
A visual comparison of the Inverted Bottleneck (IBN) \cite{mobilenetv2} and Fused-IBN \cite{mobiledet,nahas,efficientnetv2} blocks.
}
\label{fig:ibn_v_fused}
\vspace{-1em}
\end{figure}

%% file: fig/qualitative_cityscapes.tex
\newcommand{\imgsizeC}{0.97}

\begin{figure*}[!t]
\begin{center}
    \begin{subfigure}[b]{\imgsizeC\linewidth}
        \centering
        \captionsetup{justification=centering}
        \includegraphics[width=\linewidth]{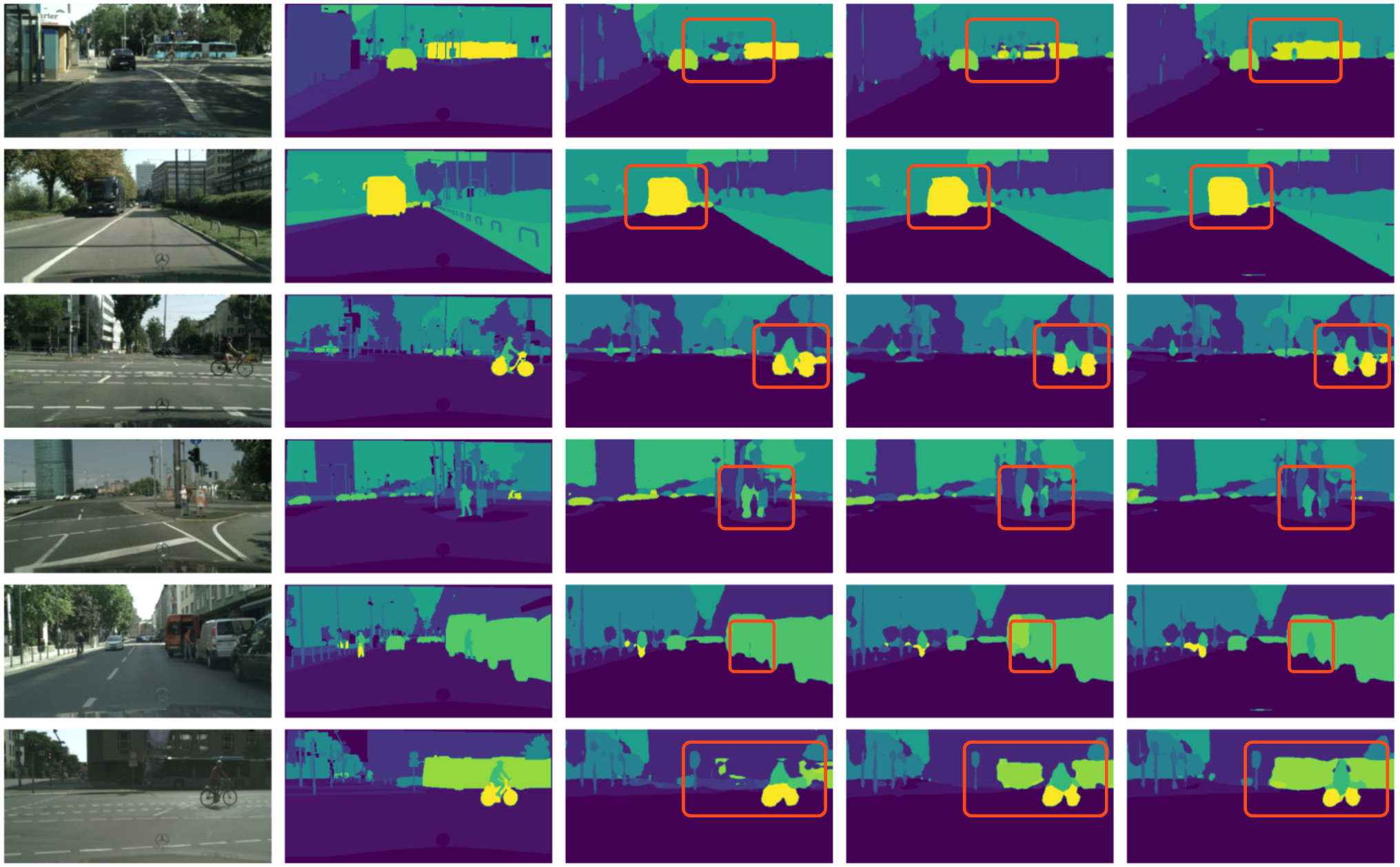}
    \end{subfigure}
    \begin{subfigure}[b]{\imgsizeC\linewidth}
        \centering
        \captionsetup{justification=centering}
        \includegraphics[width=\linewidth]{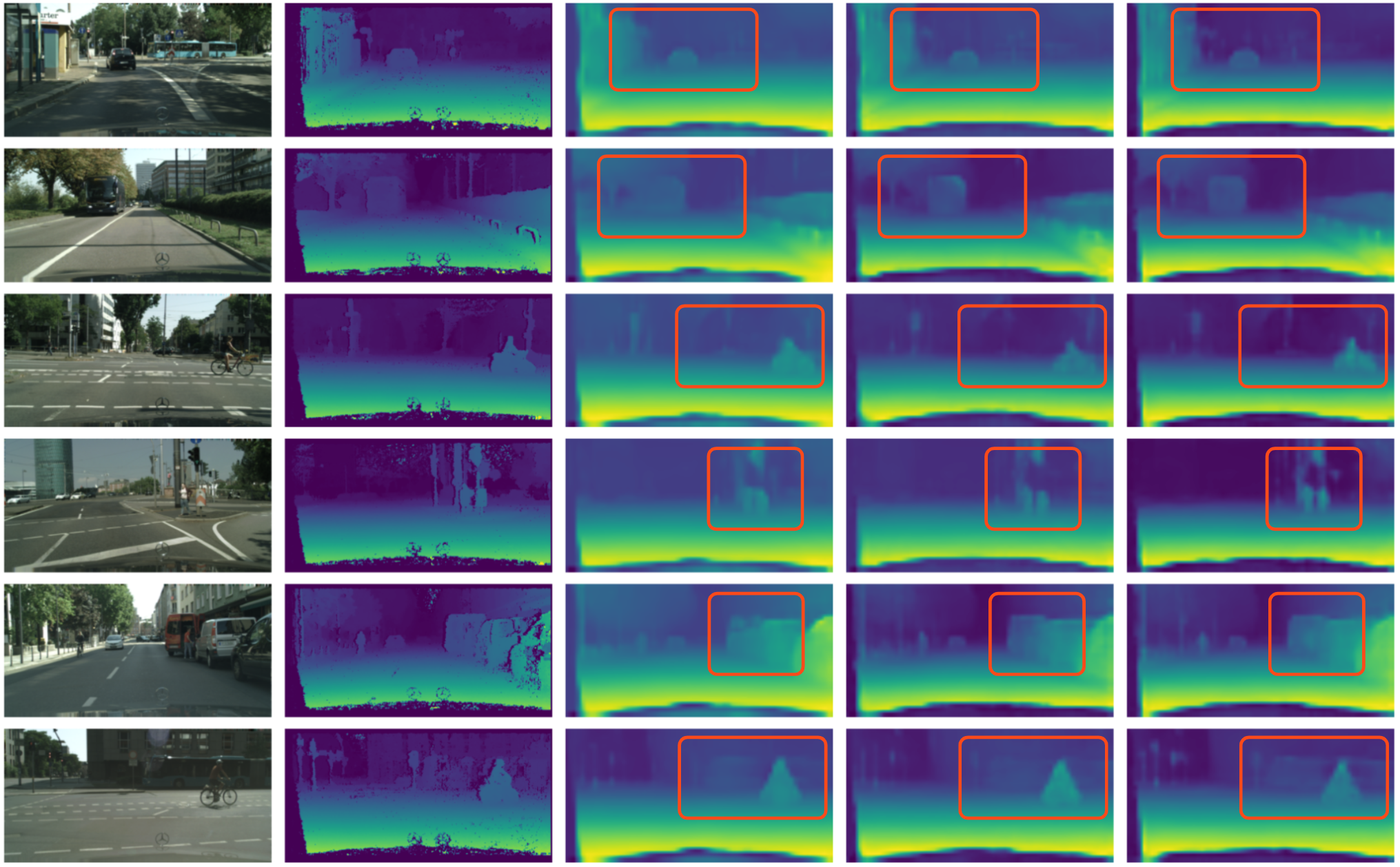}
        \caption*{
                Input Image 
            \hspace{6em} 
                Ground Truth
            \hspace{7em}  
                Single-Task 
            \hspace{7em}  
                Multi-Task
            \hspace{7.5em}  
                EDNAS
        }
    \end{subfigure}
\end{center}
\caption{{Qualitative results for semantic segmentation (top) and depth estimation (bottom) on CityScapes dataset}
}
\label{fig:qualitative_cityscapes}
\end{figure*}